\documentclass[runningheads]{llncs}
\usepackage{graphicx}
\usepackage{amsfonts,amssymb} 
\usepackage{amsmath}
\usepackage{diagbox}
\usepackage{pifont}   
\usepackage{graphicx}
\usepackage{multirow}
\usepackage{multicol}

\begin{document}
\title{Deform-Mamba Network for MRI Super-Resolution}

\author{Zexin Ji\inst{1,2,4} \and
Beiji Zou\inst{1,2} \and
Xiaoyan Kui\inst{1,2}* \and
Pierre Vera\inst{4} \and
Su Ruan\inst{3}
}

\authorrunning{Z. Ji et al.}

\institute{School of Computer Science and Engineering, Central South University,
Changsha, 410083, China \and
Hunan Engineering Research Center of Machine Vision and Intelligent
Medicine, Central South University, Changsha, 410083, China \and
University of Rouen-Normandy, LITIS - QuantIF UR 4108, F-76000, Rouen, France \and
Department of Nuclear Medicine, Henri Becquerel Cancer Center, Rouen, France\\
\email{zexin.ji@csu.edu.cn}
}

\maketitle              
\begin{abstract}

In this paper, we propose a new architecture, called Deform-Mamba, for MR image super-resolution. Unlike conventional CNN or Transformer-based super-resolution approaches which encounter challenges related to the local respective field or heavy computational cost, our approach aims to effectively explore the local and global information of images. Specifically, we develop a Deform-Mamba encoder which is composed of two branches, modulated deform block and vision Mamba block. We also design a multi-view context module in the bottleneck layer to explore the multi-view contextual content. Thanks to the extracted features of the encoder, which include content-adaptive local and efficient global information, the vision Mamba decoder finally generates high-quality MR images. Moreover, we introduce a contrastive edge loss to promote the reconstruction of edge and contrast related content.
Quantitative and qualitative experimental results indicate that our approach on IXI and fastMRI datasets achieves competitive performance.

\end{abstract}
\section{Introduction}
Magnetic resonance imaging (MRI) is one of the broadly used medical imaging techniques for disease diagnosis. The use of high-resolution MR images can allow radiologists to diagnosis accurately as it provides more anatomical and pathological information. However, high-quality MRI requires a long echo time (TE) or repetition time (TR) and needs expensive medical equipment. Obtaining high-resolution MR images from limited scanning time and acquired hardware is highly demanded in clinical. Thus, improving the resolution of MRI is particularly important. The technique of image super-resolution (SR) has drawn a lot of attention in medical image processing. Specifically, the MRI super-resolution task aims to synthesize high-resolution~(HR) MR images from the low-resolution~(LR) counterparts in a cost-effective way. 

The two main categories of super-resolution techniques are traditional and deep learning-based techniques. Bicubic and b-spine~\cite{1} are two examples of interpolation-based traditional SR techniques. 
Other more complex traditional
methods have also been developed,
such as methods based on
reconstruction~\cite{2,3}, and
methods based on examples~\cite{4,5}.
These traditional methods use their limited prior knowledge to improve image resolution, which has fast processing speed but always blurs details. Benefiting from the strong feature representation ability of convolutional neural networks (CNNs), deep learning-based methods have been used for super-resolution task~\cite{8,9,10,11,12}
Super-resolution convolutional neural network~(SRCNN)~\cite{8} firstly applied CNNs to the SR task. Later, the further improvements are all based on the SRCNN~\cite{9,10,11,13}. For example, Qiu \emph{et al.}~\cite{10} adopted SRCNN~\cite{8} and a sub-pixel convolutional layer to obtain high-quality MR images. Furthermore, a feedback adaptive weighted dense network (FAWDN)~\cite{11} was developed for high-resolution medical image reconstruction.
However, these methods only exploit the knowledge of a limited receptive field, ignoring the long-range dependencies in the image. To solve this issue, vision Transformer has emerged in the low-level tasks~\cite{liang2021swinir,fang2022cross,DBLP:conf/cvpr/ChenWZ0D23,25,DBLP:journals/cbm/WangSCX23}, demonstrating the ability to capture the global contexts. 
For example, Forigua \emph{et al.}~\cite{forigua2022superformer} proposed SuperFormer, which uses the swim transformer~\cite{liu2021swin} for MR image SR.
Fang \emph{et al.}~\cite{fang2022cross} designed a high-frequency
Transformer that incorporates the high-frequency structure
prior to implement image SR. Despite their remarkable success, these approaches still encounter several problems. 
(1) Calculating the self-attention in Transformers incurs space and time resources that increase quadratically as the number of tokens grows.
(2) The interaction of the range of utilized features has not been fully explored.
(3) The training strategies usually focus on pixel level differences, making it difficult to constrain the reconstruction of high-frequency and overall image distribution information.

To address the aforementioned problems, we propose a multi-scale network called Deform-Mamba for MR image super-resolution, which simultaneously incorporates the deformable locality features with computationally lightweight global complementary features into the multi-scale network.
State Space Models (SSMs)~\cite{DBLP:conf/nips/GuJGSDRR21} have been widely explored recently. 
It can linearly model 1D sequences and learn long-range dependencies. The SSMs have achieved good performance in continuous long sequence data analysis tasks such as natural language processing (NLP).
Recently, Mamba~\cite{DBLP:journals/corr/abs-2312-00752} has further improved SSMs in discrete data modeling, which is a new alternative solution for CNN or Transformer. It can further model long-range dependencies effectively with linear complexity through input-dependent selection mechanisms and hardware-aware algorithms. Some studies have explored the application of Mamba in computer vision tasks. For example, 
Zhu \emph{et al.}~\cite{zhu2024vision} designed a vision Mamba backbone with bidirectional Mamba blocks, which demonstrates its effectiveness in semantic segmentation, classification, and object detection tasks. Xing \emph{et al.}~\cite{xing2024segmamba} proposed SegMamba for 3D medical image segmentation. 
Inspired by the success of Mamba, we first propose to use Mamba to realize the MR image super-resolution task.
Different from the traditional method combining ordinary convolution and Transformer to learn the local and global features, we design a Deform-Mamba module as the basic unit to merge the strengths of both deformable CNNs~\cite{DBLP:conf/iccv/DaiQXLZHW17} and Mamba. This module can fully explore local and global features of the image according to the image content. We combine this module with the Unet network to further learn multi-scale local and global information.
Furthermore, we also develop a multi-view context module to strengthen the ability to understand image semantic content in the bottleneck layer.
To further enhance the high-frequency information of super-resolved images, we introduce a contrastive edge loss (CELoss) that focuses more on the edge texture and contrast of MR images.
The quantitative and qualitative experimental results demonstrate the effectiveness of our Deform-Mamba.

\section{Method}\label{Method}

\subsection{Preliminaries}

State Space Models (SSMs)~\cite{DBLP:conf/nips/GuJGSDRR21}, i.e., Mamba~\cite{DBLP:journals/corr/abs-2312-00752} that core is selective scan space state sequential model (S6), maps a 1-D function or sequence $x(t) \in \mathbb{R}$ to the output $ y(t) \in \mathbb{R}$ through
a hidden state $h(t) \in \mathbb{R}^{\mathbb{N}}$. It is usually calculated by the linear ordinary differential equations (ODEs). 

\begin{equation}
    h^{\prime}(t)=\mathbf{A} h(t)+\mathbf{B} x(t), y(t)=\mathbf{C} h(t),
\end{equation}
where $\mathbf{A} \in \mathbb{R}^{\mathrm{N} \times \mathrm{N}}$ is the state matrix. $\mathbf{B} \in \mathbb{R}^{\mathbb{N} \times 1}$ and $ \mathbf{C} \in \mathbb{R}^{1 \times \mathbb{N}}$ are the projection parameters.

Mamba uses the discrete rule zero-order hold $(\mathrm{ZOH})$ to discretize ODEs into discrete functions, which is more suitable for deep learning scenarios. Specifically, we use timescale parameter $\Delta$ to transform $\mathbf{A}$ and $\mathbf{B}$ into discrete parameters $\overline{\mathbf{A}}$ and $\overline{\mathbf{B}}$, respectively. The specific implementation is as follows:

\begin{equation}\label{eq2}
\overline{\mathbf{A}}=\exp (\Delta \mathbf{A}), \overline{\mathbf{B}}=(\Delta \mathbf{A})^{-1}(\exp (\Delta \mathbf{A})-\mathbf{I}) \cdot \Delta \mathbf{B} .
\end{equation}

$\mathrm{Eq}$. (1) will be the following form after the discretization:

\begin{equation}\label{eq3}
h_t=\overline{\mathbf{A}} h_{t-1}+\overline{\mathbf{B}} x_t, y_t=\mathbf{C} h_t .
\end{equation}

Finally, the model generates the output via  a global convolution as follows:
\begin{equation}\label{eq4}
\overline{\mathbf{K}}=\left(\mathbf{C} \overline{\mathbf{B}}, \mathbf{C} \overline{\mathbf{AB}}, \ldots, \mathbf{C} \overline{\mathbf{A}}^{{M}-1} \overline{\mathbf{B}}\right), \mathbf{y}=\mathbf{x} * \overline{\mathbf{K}},
\end{equation}
where $M$ denotes the length of the input sequence $\mathbf{x}$. $\overline{\mathbf{K}} \in \mathbb{R}^{\mathrm{M}}$ represents a structured convolutional kernel. This model can selectively and linearly learn the long-range dependencies of sequences.

\begin{figure}[t]
    \centering
    \includegraphics[width=1\linewidth]{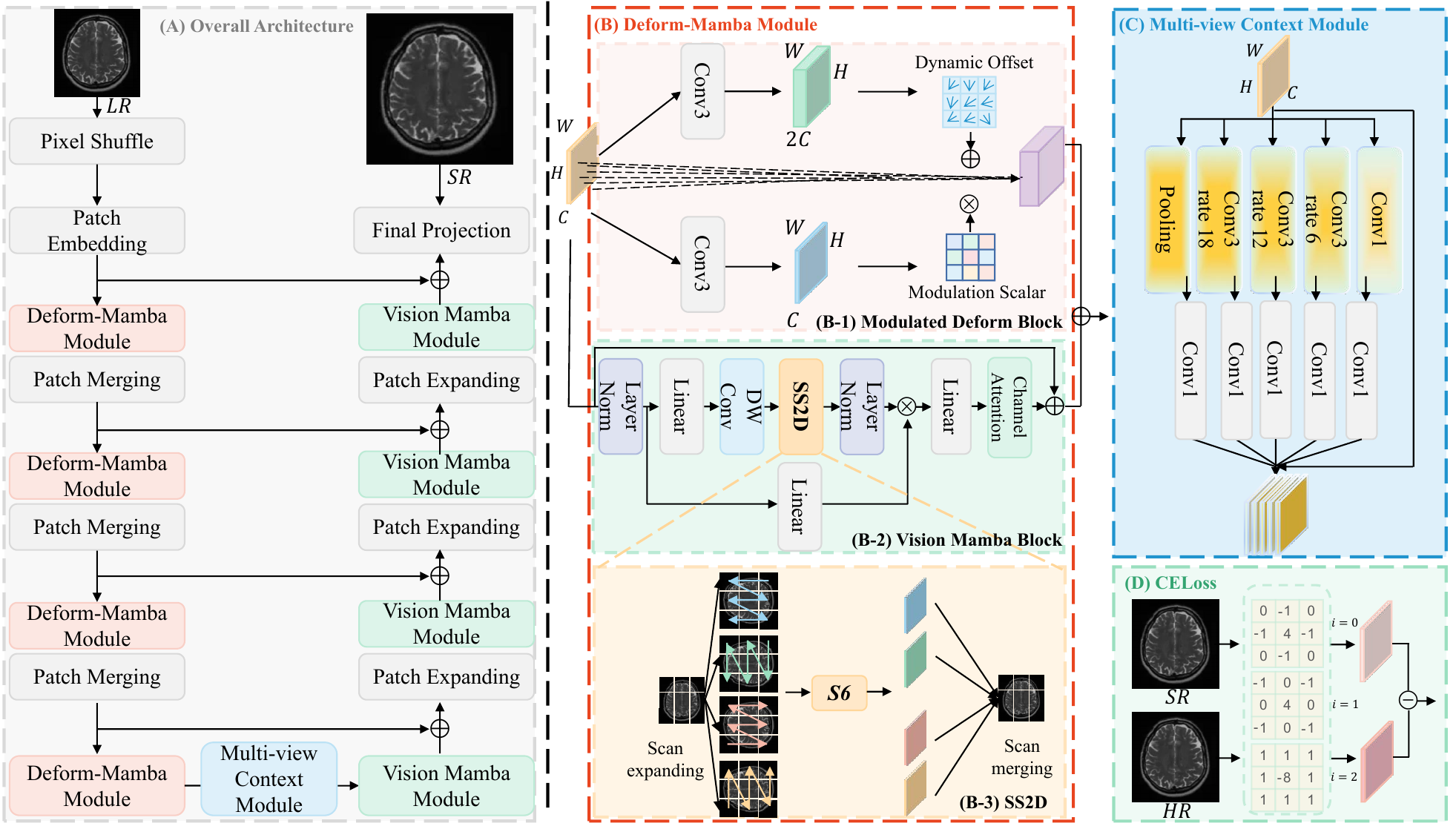}
    \caption{Architecture of our Deform-Mamba. (A) The overall architecture mainly consists of a Deform-Mamba encoder, a multi-view context module, and a vision Mamba decoder. (B) An implementation of the Deform-Mamba module includes (B-1) modulated deform block, (B-2) vision Mamba block, and (B-3) 2D-Selective-Scan (SS2D) block. (C) Multi-view context block. (D) Contrastive edge loss (CELoss).}
    \label{fig:enter-label}
\end{figure}

\subsection{Proposed Architecture}

We show the main network architecture in Fig.\ref{fig:enter-label} (A). In the first step, we use pixel shuffle to upsample low-resolution image in a learnable manner, so as not to lose image information. Our approach mainly contains patch embedding, a Deform-Mamba encoder, a multi-view context module, a vision Mamba decoder, and a final projection layer.
Specifically, the image is first cropped into non-overlapping patches through patch embedding. The encoder consists of a Deform-Mamba module and patch merging that increases the number of channels. We also design the multi-view context module to further enhance feature representation in the bottleneck layer. In the decoder, each layer consists of a pure vision Mamba module and patch expanding that reduces the number of channels. The super-resolved image is obtained through the final projection layer. Furthermore, we develop a contrastive edge loss (CELoss) to regulate the generation of edges and contrast of the image.

\noindent\textbf{Deform-Mamba Module.}
As illustrated in Fig.\ref{fig:enter-label} (B), we develop a Deform-Mamba module to extract local and global features that can be adaptive to image content. 
It mainly consists of two branches, modulated deform block and vision Mamba block.
For local operation, we use the modulated deform block to adaptively modify
the size of the receptive field of local operation, while for global operation, we apply a novel vision Mamba block for building long-distance dependence between features. 
Then we add output features of two branches to gradually fuse the multi-scale deformable local and efficient global information.

Specifically, Fig.\ref{fig:enter-label} (B-1) shows the network structure of the modulated deform block.
Different from traditional convolution using fixed shape sampling, the modulated deform block can dynamically learn the convolution bias according to the image content to adaptively extract features. The operation is characterized as follows:

\begin{equation}
Y(p)=\sum_{k=1}^K w_k \cdot X\left(p+p_k+\Delta p_k\right) \cdot \Delta m_k,
\end{equation}
where $k$ is the index.  $w_k$ denotes the $k$-th kernel weight. $X$ and $Y$ are the input and output feature map. $p$, $p_k$, $\triangle p_k$ and $\triangle m_k$ are the sampling position, $k$-th predefined offset, dynamic offset, and modulation scalar for the $k$-th location, respectively.
Specifically, $\Delta p_k \in \mathbb{R}^{H \times W \times 2 C}$ and $\Delta m_k \in \mathbb{R}^{H \times W \times C}$ are learned by the convoluional layer. $\Delta p_k=\operatorname{Conv}\left(\mathbf{X}\right)$, and $\Delta m_k=\sigma\left(\operatorname{Conv}\left(\mathbf{X}\right)\right)$. The $\sigma$ is the sigmoid function. We then add the learned $\triangle p_k$ to $p_k$ and adaptively change the sampling position based on the image content. In addition, $\triangle m_k$ can control the weight of the offset sampling points, which reduces the interference of irrelevant features.

Fig.\ref{fig:enter-label} (B-2) illustrates the architecture of the vision Mamba block. Unlike the Transformer, which employs a self-attention mechanism to calculate long-range dependencies of images. The vision Mamba block models in a sequential manner, which is more efficient than Transformer, in particular for processing high-resolution images. The input feature map first passes through layer normalization, then it diverges into dual pathways. The first pathway involves the input through a linear layer and then an activation function. In the second pathway, it is processed via a linear layer, depthwise separable convolution, and an activation function, subsequently into the 2D-Selective-Scan (SS2D) and layer normalization. Then we use multiplication to integrate both pathways. We also utilize channel attention to model the correlation between different channels.
Fig.\ref{fig:enter-label} (B-3) shows the specific implementation of SS2D. The scan expanding first unfolds the image in four different directions to fully mine effective information. 
The main core module S6 of Mamba is used to connect with previous patches of the image by the hidden state space to lightly learn long-range dependencies (Eq.\ref{eq2}-Eq.\ref{eq4}). For more details of S6, please refer to \cite{DBLP:conf/nips/GuJGSDRR21}. Finally, the scan merging can merge four features of different sequences to recover the original size.

\noindent\textbf{Multi-view Context Module.}
Inspired by~\cite{DBLP:journals/pami/ChenPKMY18},
we also develop a multi-view context module in the bottleneck layer to improve the ability of feature extraction by 
exploring the multi-view information. The bottleneck layer is a crucial location for feature extraction and information aggregation.
As shown in Fig.\ref{fig:enter-label} (C), we parallelly utilize atrous convolution with different dilation rates to increase the receptive field of the convolution kernel.
These features are concatenated to fuse feature information at different scales, without increasing the number of parameters and computational complexity. Then we use the residual connection to 
get the output of the bottleneck layer.

\noindent\textbf{Loss Function.}
We utilize the $\mathcal{L}_1$ loss function to measure the difference between the network prediction and the ground truth image. 
However, pixel-level loss functions primarily concentrate on the individual disparities among pixels, failing to comprehensively grasp the structural intricacies within an image. Therefore, we design a contrastive edge loss (CELoss) to constrain the generation of more detailed super-resolved MR images by enhancing the edge and local contrast information.

\begin{equation}
\mathcal{L}_{CELoss}=\sum_i\left\|\mathbf{E}_i \odot SR-\mathbf{E}_i \odot HR \right\|_2^2,
\end{equation}
where $SR$ is the super-resolved MR image, and $HR$ is the ground-truth high-resolution image. $\mathrm{E}_i$ is the $i$ th contrastive edge convolution kernel. As shown in Fig.\ref{fig:enter-label} (D), $i \in$ $[0,2]$. 
Specifically, the three kernels are designed to emphasize the edges of the image, as well as local contrast, to highlight the details of liquid and moisture regions in MR images.
The final loss function ensures that the SR model not only reconstructs the
image at the pixel level but also restores critical visual features about edge and contrast. It is represented as:

\begin{equation}
\begin{split}
Loss = \mathcal{L}_{1} + \beta \mathcal{L}_{CELoss}, 
\end{split}
\end{equation}
where the weight $\beta$ for $\mathcal{L}_{CELoss}$ is 0.1 in our experiment.

\section{Experiments}\label{Experiments}
\subsection{Datasets and Evaluation Metrics}
To evaluate the effectiveness of our approach, we conducted experiments on the IXI \footnote{http://brain-development.org/ixi-dataset/} and fastMRI \footnote{https://fastmri.med.nyu.edu/} with the brain and knee T2 weighted image. The slice resolutions of IXI and fastMRI are 256$\times$256 and 320$\times$320, respectively. We utilized 368 subjects from the IXI dataset for training and 92 for testing. For the fastMRI dataset, 227 subjects were used for training and 45 for testing. To synthesize low-resolution input images, we implement it in the frequency domain~\cite{36} to further fit the true distribution of low-resolution images. We utilized the peak signal to noise ratio (PSNR) and structural similarity index (SSIM)~\cite{45} to evaluate the quality of super-resolved images.

\subsection{Experimental Details}
We implemented our approach with the Pytorch toolbox and trained the network on the NVIDIA RTX A6000 GPU. We used Adam optimizer with the initial learning rate of $1 \times 10^{-4}$ to update network parameters. The network was trained for 50 epochs with the batch size of 2. We used 4 vision Mamba blocks in each level. The channel count of each level is [96,128,384,768].


\subsection{Ablation Study}

We studied the importance of the designed components in our Deform-Mamba. To validate the effect of deformable local features for super-resolution, we remove the modulated deform block. It represented \textit{w/o Deform}, which is the pure vision Mamba network. Further, we remove the multi-view context module to evaluate the effect of multi-scale contextual information in the bottleneck layer. This model is called \textit{w/o MVC}. We also verify the effect of contrastive edge loss, which is denoted as \textit{w/o CELoss}. The overall result of our model is \textit{Deform-Mamba}. From the Tab.\ref{tab1} we can see that the results of \textit{w/o Deform}, \textit{w/o MVC}, and \textit{w/o CELoss} are worse than our \textit{Deform-Mamba}. It indicates the necessity of deformable local features and multi-view contextual information for super-resolution, and the constraint of edge and contrast is useful to generate high-frequency information.

\begin{table}[t]
\centering
\caption{Ablation study with different components in our Deform-Mamba}
\setlength{\tabcolsep}{7mm}{
\begin{tabular}{c|cc}
\hline
Method & PSNR$\uparrow$  & SSIM$\uparrow$  \\ \hline
\textit{w/o Deform}      & 32.42    & 0.9250    \\
\textit{w/o MVC}      & 32.60    & 0.9266    \\
\textit{w/o CELoss}      & 32.59    & 0.9264    \\
\textit{Deform-Mamba}      & \textbf{32.65}    & \textbf{0.9270}    \\ \hline
\end{tabular}}
\label{tab1}
\end{table}

\begin{table}[t]
\centering
\caption{Quantitative results with different methods under fastMRI and IXI dataset}
\setlength{\tabcolsep}{5mm}{
\begin{tabular}{c|cc|cc}
\hline
\multirow{2}{*}{Method} & \multicolumn{2}{c|}{fastMRI 4$\times$} & \multicolumn{2}{c}{IXI 4$\times$} \\  
                        & PSNR$\uparrow$           & SSIM$\uparrow$           & PSNR$\uparrow$          & SSIM$\uparrow$        \\ \hline
SRCNN                   & 19.74          & 0.3653       & 28.12        & 0.8357     \\
VDSR                    & 20.31          & 0.3839        & 28.34        & 0.8392     \\
FMISR                   & 24.35          & 0.5207        & 28.27      & 0.8349     \\
T$^{2}$Net                   & \underline{30.56}           & \underline{0.6244}         & 29.73      & 0.8773     \\
HAT                     & 29.65        & 0.6155        & \textbf{30.73}      & \textbf{0.9007}      \\
Our                     & \textbf{32.11}        & \textbf{0.7194}        & \underline{30.60}       & \underline{0.8965}      \\ \hline
\end{tabular}}
\label{tab2}
\end{table}

\begin{figure}[t]
    \centering
    \includegraphics[width=1\linewidth]{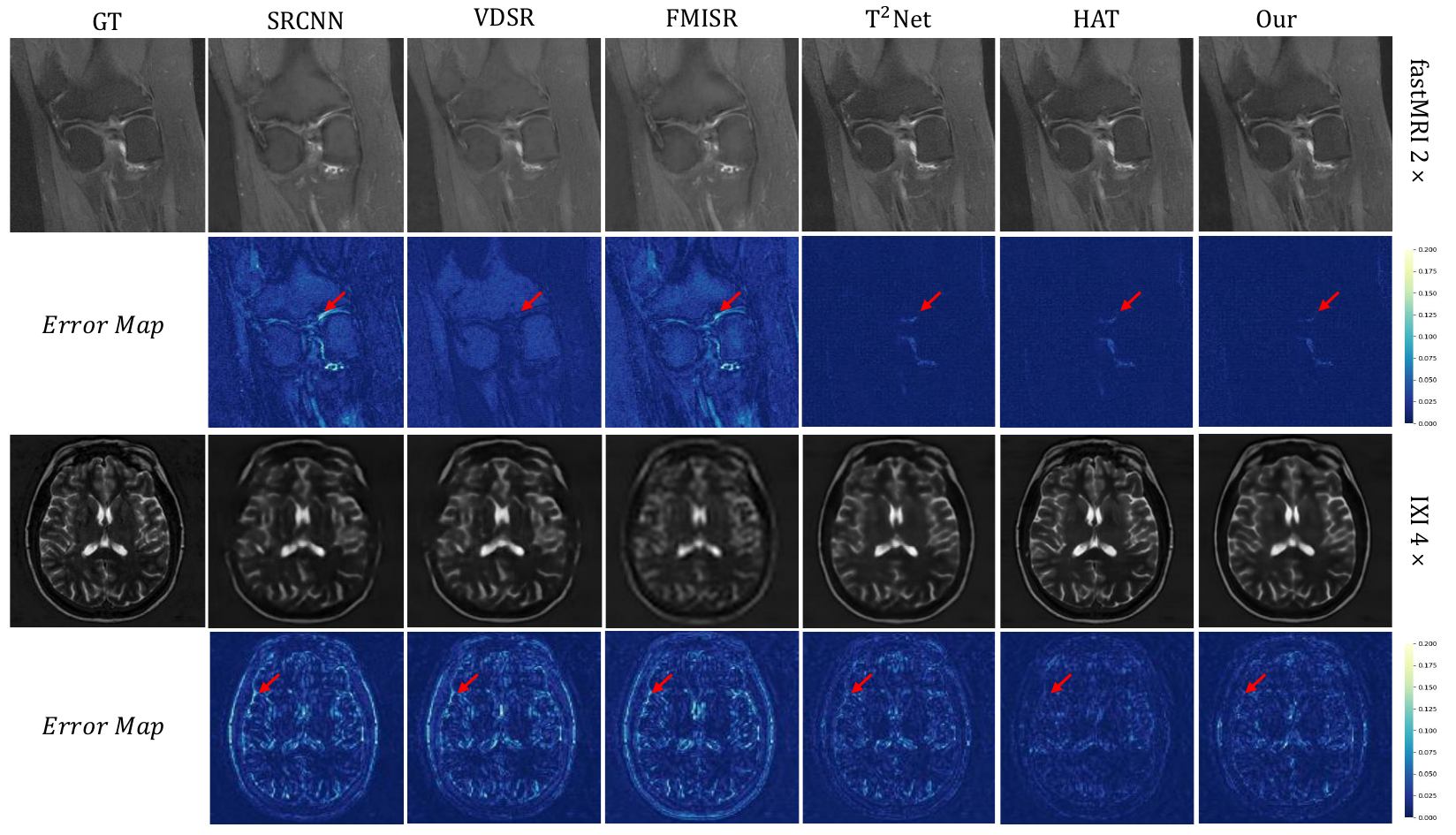}
    \caption{Qualitative results with different methods under fastMRI and IXI dataset}
    \label{fig:VisualResult}
\end{figure}

\subsection{Comparison with state-of-the-art methods}

\textbf{Quantitative Analysis.}
For quantitative evaluation, we compared the PSNR and SSIM values of SRCNN~\cite{8}, VDSR~\cite{9}, FMISR~\cite{19},
T$^2$Net~\cite{12}, HAT~\cite{DBLP:conf/cvpr/ChenWZ0D23} and our approach on the IXI and fastMRI dataset under $4 \times$ upsampling factors. We include the results of the $2 \times$ upsampling in the supplementary materials. The quantitative comparison experiment results are shown in Tab.\ref{tab2}. It can be seen from the table that the PSNR and SSIM values of our approach on the fastMRI dataset obtain the best performance than those of other methods. This is because our proposed Deform-Mamba model can adaptively mine the local and global content of images. 
For the IXI dataset, we achieved the second-best results compared to transformer-based HAT. Our method using Mamba with linear computational complexity takes less time to train as fewer parameters used in our architecture.

\noindent\textbf{Qualitative Analysis.}
The super-resolved results and corresponding error maps are shown in Fig.\ref{fig:VisualResult} under the fastMRI and IXI datasets. 
The error map is typically a visualization of the difference between the super-resolved image and the ground truth high-resolution image. The darker the color, the smaller the difference between the generated image and the label image, and vice versa. The figures show that the compared methods on the fastMRI dataset produce blurring artifacts since the available information in the image is not fully utilized by these approaches. It can be clearly seen in the error map that the color of our method is the darkest.
For the IXI dataset, our method with fewer parameters achieved the best local visualization results.

\section{Conclusion and Future Work}\label{Conclusion}
In this paper, we have developed a Deform-Mamba network for MR image super-resolution. Our model combines a modulated deform block and vision Mamba block as a unit in the encoder, which can activate more content-adaptive local and global features for super-resolution. The designed multi-view context block in the bottleneck layer can further enhance the fusion of multi-scale contextual information. The contrastive edge loss further reconstructs edge-enhanced and contrast-consistent high-resolution images. Quantitative and qualitative experiments demonstrate the effectiveness of our approach. In future work, our Deform-Mamba network can be added as a baseline to the diffusion model to further improve the performance of super-resolution networks.

\noindent\textbf{Acknowledgements.} The work was supported by the National Key R$\&$D Program of China (No.2018AAA0102100); the National Natural Science Foundation of China (Nos.U22A2034, 62177047); the Key Research and Development Program of Hunan Province (No.2022SK2054); Major Program from Xiangjiang Laboratory under Grant 23XJ02005; Central South University Research Programme of Advanced Interdisciplinary Studies (No.2023QYJC020); the Natural Science Foundation of Hunan Province (No.2024JJ6338); the Fundamental Research Funds for the Central Universities of Central South University (No.2024ZZTS0486); the China Scholarship Council (No.202306370195)

\bibliographystyle{splncs04}
\bibliography{mybibfile}






\end{document}